\documentclass[10pt,twocolumn,letterpaper]{article}
\pdfoutput=1 
\usepackage{cvpr}
\usepackage{times}
\usepackage{epsfig}
\usepackage{graphicx}
\usepackage{amsmath}
\usepackage{amssymb}
\usepackage{subfigure}

\usepackage{paralist}

\usepackage[pagebackref=false,letterpaper=true,colorlinks,bookmarks=false]{hyperref}

\cvprfinalcopy 

\def\cvprPaperID{1719} 

\newcommand{\figref}[1]{Figure~\ref{#1}}
\newcommand{\secref}[1]{Sec.~\ref{#1}}

\newcommand{\refequ}[1] {Eq.~(\ref{#1})}

\newcommand{\reftab}[1] {Table~\ref{#1}}

\ifcvprfinal\fi
\begin{document}

\title{Thin-Slicing Network: A Deep Structured Model for Pose Estimation in Videos}

\author{Jie Song$^{1}$ \quad \quad Limin Wang$^{2}$ \quad \quad Luc Van Gool$^{2}$ \quad \quad Otmar Hilliges$^1$ \\
\small $^{1}$Advanced Interactive Technologies, ETH Zurich \quad \quad
\small $^{2}$Computer Vision Laboratory, ETH Zurich
}

\maketitle

\begin{abstract}
Deep ConvNets have been shown to be effective for the task of human pose estimation from single images. However, several challenging issues arise in the video-based case such as self-occlusion, motion blur, and uncommon poses with few or no examples in training data sets. Temporal information can provide additional cues about the location of body joints and help to alleviate these issues.
In this paper, we propose a deep structured model to estimate a sequence of human poses in unconstrained videos.
This model can be efficiently trained in an end-to-end manner and is capable of representing appearance of body joints and their spatio-temporal relationships simultaneously. Domain knowledge about the human body is explicitly incorporated into the network providing
effective priors to regularize the skeletal structure and
to enforce temporal consistency. The proposed end-to-end architecture is evaluated on two widely used benchmarks (Penn Action dataset and JHMDB dataset) for video-based pose estimation. Our approach significantly outperforms the existing state-of-the-art methods.
\end{abstract}


\section{Introduction}
\label{sec:introduction}
\begin{figure}[ht]
\begin{center}
   \includegraphics[width=1\linewidth]{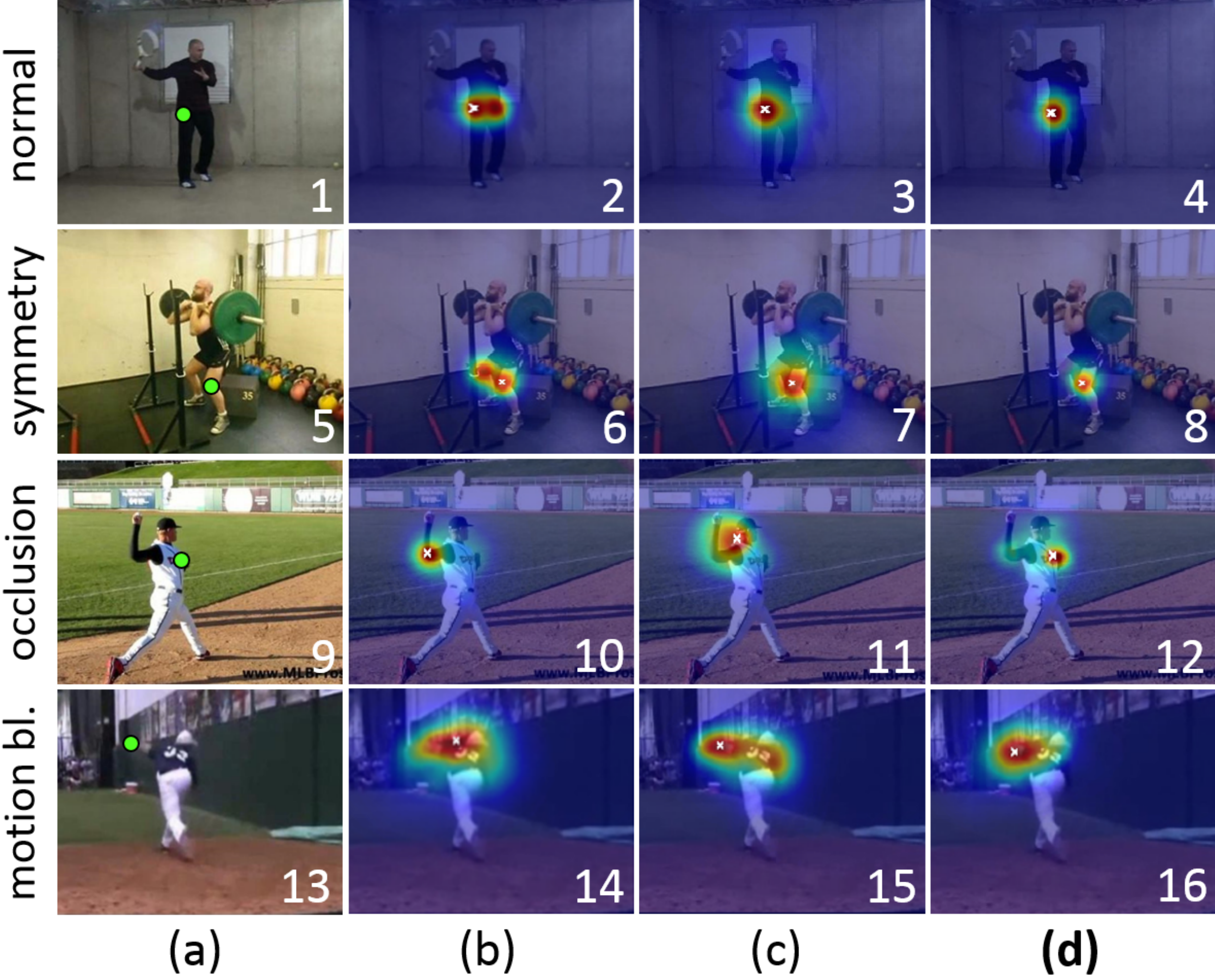}
\end{center}
   \caption{Our method incorporates spatio-temporal information into a single end-to-end trainable network architecture, aiming to deal with challenging problems such as (self-)occlusions, motion blur, and uncommon poses.
Taking fully unconstraind images as input (a), we regress body-part locations with standard ConvNet layers (b). Spatial inference helps in overcoming confusion due to symmetric body parts (c). Our spatio-temporal inference layer (d) can deal with extreme cases where spatial information only fails (cf. 11 vs 12, 15 vs 16) \emph{and} improves prediction accuracies for unary terms due to repeating measurements by temporal propagation of joint position estimates (3 vs 4).}
\label{fig:long}
\label{fig:motivation}
\end{figure}

Estimating human poses is one of the core problems in computer vision and has many applications in the life-sciences, computer animation and the growing fields of robotics, augmented and virtual reality. Accurate pose estimates can also drastically improve the performance
of activity recognition and high-level analysis of videos(cf. ~\cite{Jhuang:ICCV:2013,wang2014video,xiaohan2015joint}). Recent pose estimation methods have exploited deep convolutional networks (ConvNets) for body-part detection in single, fully unconstrained images~\cite{chen2014articulated,newell2016stacked,ouyang2014multi,pishchulin2015deepcut,tompson2014joint,toshev2014deeppose,wei2016cpm}. While demonstrating the feasibility of detection-based pose estimation from images taken under general conditions, such methods still struggle with several challenging aspects including the diversity of human appearance and self-symmetries. Several methods~\cite{chen2014articulated,yang2016end} have explicitly incorporated geometric constraints among body parts into such frameworks, ensuring spatial consistency and penalizing physically impossible solutions (cf. \figref{fig:motivation}, (c)).

In this paper we consider the comparatively less studied problem of human pose estimation from unconstrained \emph{videos}~\cite{gkioxari2016chained,pfister2015flowing,zhang2015human,zuffi2013estimating}. While inheriting many properties from image-based pose estimation, it also brings new challenges. In particular, unconstrained videos such as those found in online portals, contain many frames with occlusions, unusual poses, and motion blur (see \figref{fig:motivation}). These issues continue to limit the accuracy of joint detection even if taking priors about the spatial configuration of the human skeleton into consideration, and often result in visible jitter if such models are applied directly to video sequences.

To tackle these problems, we propose to incorporate spatial and temporal modeling into deep learning architectures.
The proposed model is based on a simple observation: human motion exhibits high temporal consistency, which could be captured by optical flow warping~\cite{pfister2015flowing,zhang2015human,zuffi2013estimating} and spatio-temporal inference~\cite{wang2014video,xiaohan2015joint}. Specifically, we leverage a spatio-temporal relational model into the ConvNet framework and develop a new deep structured architecture, called {\em Thin-Slicing Network}. 
Our deep structured model allows for end-to-end training of body part regressors and spatio-temporal relational models in a unified framework. This enables improving generalization performance by regularizing the learning process both spatially and temporally across adjacent frames.
We deploy fully ConvNet for initial part detection. Furthermore, via a flow warping layer which propagates joint prediction heat
maps temporally and a novel inference layer, message passing
on arbitrary loopy graphs along both spatial and temporal
edges is performed.


In consequence, our approach can deal with many challenging situations arising in unconstrained video, and outperform both the original joint-position estimation methods and those incorporating spatial priors only. \figref{fig:motivation}  illustrates how our approach can accurately predict joint positions in difficult situations of full occlusion (3\textsuperscript{rd} row, given visibility in adjacent frames) or severe motion blur (4\textsuperscript{th} row, by exploiting temporal consistency). Last but not least, the model also improves predictions in relatively simple cases (see \figref{fig:motivation}, 1\textsuperscript{st} and 2\textsuperscript{nd} row). This can be explained by optimizing of several correlated but different frames through the entire architecture jointly, which not only learns weights of the inference layers, but also refines the underlying ConvNet-based part regressors, resulting in more accurate joint detections.


In summary our main contributions are:
\begin{inparaenum}[(i)]
\item A structured model captures the inherent consistency of human poses in video sequences based on a loopy spatio-temporal graph. Our approach does not rely on explicit human motion priors but leverages dense optical flow to exploit image evidence from adjacent frames.
\item An efficient and flexible inference layer performs message passing along the spatial and temporal graph edges and significantly reduces joint position uncertainty.
\item The entire architecture well integrates the ConvNet-based joint regressors and the high-level structured inference model in a unified framework, which could be optimized in an end-to-end manner. 
\item Our method significantly improves the state-of-the-art performance on two widely used video based pose estimation benchmarks: the Penn Action dataset~\cite{Zhang2013From}, the JHMDB dataset~\cite{Jhuang:ICCV:2013}.
\end{inparaenum}

%
%
%
%



\begin{figure*}
\begin{center}
   \includegraphics[width=0.9\linewidth]{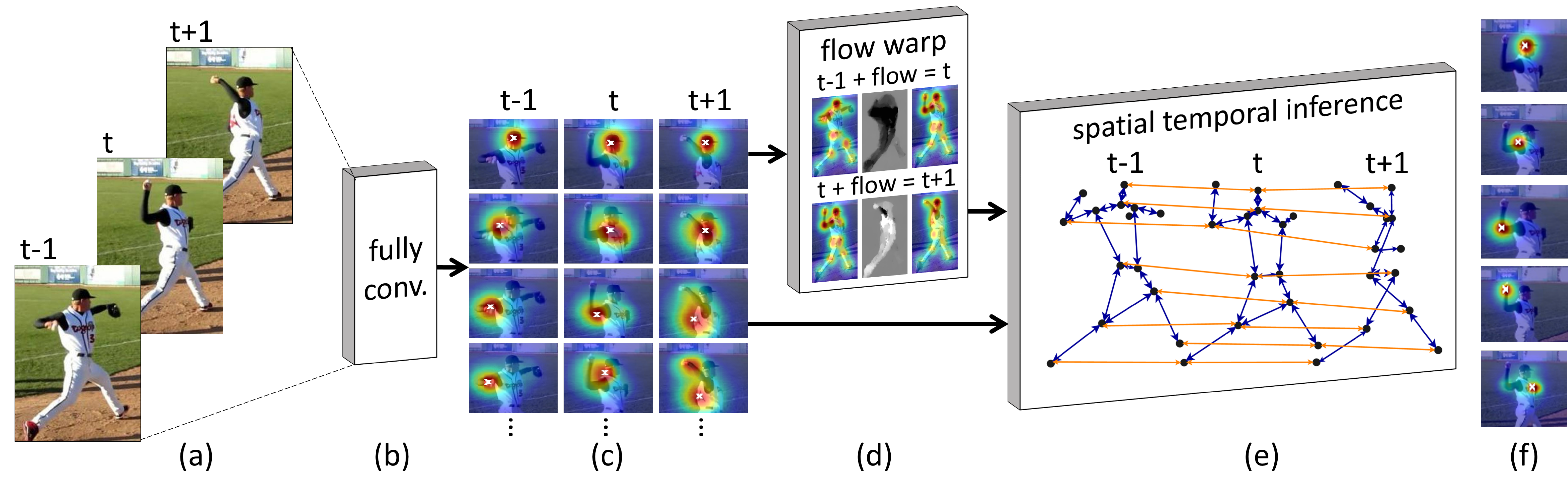}
\end{center}
   \caption{\textbf{Schematic overview of Thin-Slicing Network architecture}. Our model takes a small number of adjacent frames as input (a) and fully convolutional layers (b) regress initial body joint position estimates (c). We compute dense optical flow between neighboring frames to propagate joint position estimates through time. A flow based warping layer aligns joint heat-maps to the current frame (d). A spatio-temporal inference layer performs iterative message passing along both spatial and temporal edges of the loopy pose configuration graph (e) and computes final joint position estimates (f).}
\label{fig:pipeline}
\end{figure*}



\section{Related work}
\label{sec:rw}
Pose estimation from \textbf{single images} has benefitted tremendously from leveraging structural models such as tree-structured pictorial models~\cite{andriluka2009pictorial} and part-based models~\cite{johnson2010clustered,pishchulin2013poselet,ramanan2006learning,yang2011articulated}, encoding the relationships between articulated joints. While capturing kinematic correlations, such models are prone to errors such as double-counting part evidence.
More expressive loopy graph models, allowing for cyclic joint dependencies have been proposed to better capture symmetry and long-range correlation~\cite{dantone2013human,ren2005recovering,sigal2006measure,sun2011articulated}. Since exact inference in cyclic graphs is generally speaking intractable, approximate inference methods like loopy belief propagation are typically used. 

The above methods are based on hand-crafted features and are sensitive to (the limits of) their representative power. More recently, \textbf{convolutional deep learning} architectures have been deployed to learn richer and more expressive features directly from data~\cite{chen2014articulated,ouyang2014multi,tompson2014joint,pishchulin2015deepcut,toshev2014deeppose}, outperforming prior work. Toshev et al.~\cite{toshev2014deeppose} directly regress the joint coordinates from images. Follow-up work suggests that regressing full image confidence maps as intermediate representation can be more effective~\cite{tompson2014joint,chen2014articulated}. While multi-stage convolutional operations can capture information in large receptive fields, they still lack the ability to fully model skeletal structure in their predictions.

Several approaches to refine confidence maps have been proposed.
First, additional convolutional layers taking joint heat-maps as input can be added to learn implicit spatial dependencies without requiring explicit articulated human body priors~\cite{tompson2014joint,wei2016cpm,chu2016structure}.
Second, \cite{pishchulin2015deepcut,chen2014articulated} explicitly resort to graphical models to post-process regressed confidence maps.
However, the parameters of part regression networks and spatial inference are learned independently in ~\cite{chen2014articulated,pishchulin2015deepcut}. In~\cite{yang2016end} an end-to-end trainable framework, combining convolutional operations and spatial refinement is proposed. Our work not only incorporates spatial information but also models temporal dependencies.

\textbf{Pose estimation in videos} brings new challenges (illustrated in \figref{fig:motivation}) and requires the coupling of parts across frames to ensure accurate and temporally stable predictions.
Early work initializes a temporal tracker from few predicted poses in the sequence's initial frames ~\cite{sidenbladh2000stochastic} but suffers from pose drift. Tracking-by-detection schemes have been used to more robustly estimate poses in videos~\cite{fragkiadaki2013pose,park2011n,ramanan2005strike}.
Researchers have also attempted to design spatio-temporal graphs to capture motion in short video sequences~\cite{cherian2014mixing,ferrari2008progressive,lee2009human,sapp2011parsing,sminchisescu2003estimating,wang2008gaussian,zhang2015human,wang2014video,xiaohan2015joint}. However, modeling spatial and temporal dependencies explicitly results in highly inter-connected models (i.e., loopy graphs with large tree-width) and exact inference becomes again intractable. One solution is to resort to
approximate inference, for instance using sampling based approaches~\cite{sminchisescu2003estimating,wang2008gaussian} or loopy belief propagation~\cite{lee2009human,ferrari2008progressive}. Alternatively, approximating the original large loopy model into one or multiple simplified tree-based models allows for efficient exact inference ~\cite{cherian2014mixing,zhang2015human}.

Some recent deep learning methods aide predictions in the current frame with information from its neighbors \cite{jain2014modeep}. Similar to our approach, \cite{pfister2015flowing} directly propagates joint position estimates from previous to the current frame via optical flow. Warped heatmaps from multiple nearby frames are combined as weighted average. Chain models~\cite{gkioxari2016chained} can capture longer temporal dependencies but makes assumptions about regular motion patterns. Our approach also incorporates spatio-temporal models into deep ConvNets but differs in that it
\begin{inparaenum}[(i)]
  \item explicitly models the spatial configuration of human poses;
  \item regularizes temporal joint positions using dense optical flow via
  \item a novel inference layer, performing message passing on general loopy spatio-temporal graphs;
  \item and is end-to-end trainable.
\end{inparaenum}



\section{Thin-Slicing Networks}
\figref{fig:pipeline} shows an overview of our proposed network architecture, consisting of several interconnected layers. Given a thin-slice of a video sequence (i.e., a small number of adjacent frames), a spatial fully ConvNet first regresses joint confidence maps (‘heat-maps’) of joint positions for each input frame (\figref{fig:pipeline}, (c)).
These heat-maps are sent into a \emph{flow warping} layer and a \emph{spatio-temporal inference} layer. The flow warping layer (\figref{fig:pipeline}, (d)) warps the body part heat-maps by pixel-wise dense optical flow tracks so that they align with its neighboring frame. Finally, both the warped heat-maps and the part heat-maps of the current frame pass through the \emph{spatio-temporal inference} layer (\figref{fig:pipeline}, (e)). This layer conducts inference between body parts spatially and temporally, producing the final joint position estimates (\figref{fig:pipeline}, (f)).

\subsection{Fully convolutional joint regression layer}
\label{subsection:fully ConvNet}
Several recent works regress heat-maps of body joints
via ConvNets~\cite{chen2014articulated,ouyang2014multi,pishchulin2015deepcut,tompson2014joint,wei2016cpm,newell2016stacked}. Such models are usually consist entirely of convolutional operations combined with spatial pooling layers. We leverage such a ConvNet~\cite{wei2016cpm} as basis for our architecture. More specifically as joint detection layers shown in \figref{fig:pipeline}, (b). Such models have already demonstrated the ability to capture local appearance properties and outperform hand-designed shallow features by large margins, but occlusions, (self-)symmetries and motion blur still pose significant challenges (cf. ~\figref{fig:motivation}). In order to alleviate these problems, a novel spatio-temporal message passing layer (\secref{subsection:message passing}) is proposed and incorporated into the network for end-to-end training.

\subsection{Flow warping layer}
\label{subsection:flow-layer}
While our goal is to improve temporal stability of joint predictions, we do not incorporate an explicit motion pattern (since human motion tends to be too unpredictable) but instead rely on dense optical flow to propagate information temporally. The joint detection heat-maps, produced by fully convolutional layers, is passed through the flow warping layer to align heat-maps from one frame to the targeted neighbor (\figref{fig:pipeline}, (d)). 
Pixel-wise flow vectors are used to align confidence estimates in neighboring frames to the target frame by shifting confidence values along the track directions. Next, these warped heat-maps serve as input to the spatio-temporal inference layer.

\subsection{Spatio-temporal inference layer}
\noindent Incorporating domain specific knowledge into deep networks has been proven to be effective in many vision tasks such as object detection~\cite{girshick2015deformable} and semantic segmentation~\cite{zheng2015conditional}. In this work, we propose to explicitly incorporate spatio-temporal dependencies into an end-to-end trainable framework.\\

\noindent\textbf{Modeling}\\
Let $G = (V,E)$ represent a graph as shown in~\figref{fig:pipeline} (e), with vertices $V$ and edges $E \subseteq  V \times V$ denoting the spatio-temporal structure of a human pose. $K = |V|$ is the number of body parts, and $i \in \{1,...,K\}$ is the $i$\textsuperscript{th} part. Each vertex corresponds to one of the body parts (i.e., head, shoulders), and each edge represents a connection between two of these parts spatially (blue arrows in~\figref{fig:pipeline}, (e)) or between the same part but distributed temporally (yellow arrows in~\figref{fig:pipeline}, (e)). We denote these edges as $E_s$ and $E_f$ respectively. Given an image $I$, a pose $p$ with respect to this graph $G$ is defined as a set of 2D coordinates in the image space representing the positions of the different body parts: $p = \{p_i = (x_i,y_i) \in  \mathbb{R}^2 : \forall i \in V\}$.
The single-image pose estimation problem then can be formulated
as the maximization of the following score $S(I,p)$ for a pose $p$ given an image $I$:
\begin{equation}
\begin{aligned}\label{equation:scorePose}
S(I,p) &= \sum_{i \in V}\phi_{i}(p_i| I) + \sum_{(i,j)\in E_s} \psi_{i,j}(p_i, p_j),\\
\end{aligned}
\end{equation}
where $\phi_{i}(p_i| I)$ is the unary term for the body part $i$ at
the position $p_i$ in image $I$ and $\psi_{i,j}(p_i, p_j)$ is the pairwise term modeling the spatial compatibility of two neighboring parts $i$ and $j$. The unary term provides confidence values of part $i$ based on the local appearance and it is modeled by the fully ConvNet (\secref{subsection:fully ConvNet}). For pairwise term we use a spring energy model to measure the deformation cost, where $\psi_{i,j}(p_i, p_j)$ is defined as $w_{i,j}\cdot d(p_i-p_j)$. With standard quadratic deformation constraints $d(p_i-p_j) = [\Delta x \quad \Delta x^2 \quad \Delta y \quad \Delta y^2]^T$, where $\Delta x = x_i - x_j$ and $\Delta y = y_i - y_j$ are the relative positions of part $i$ with respect to part $j$. The parameter $w_{i,j}$ encodes rest location and rigidity of each spring, which can be learned together with the whole network.

Given a slice of a video sequence $\mathbb{I} = (I_1,I_2,...,I_T)$ as shown in~\figref{fig:pipeline} (a), the temporal links (yellow arrows in ~\figref{fig:pipeline}, (e)) are introduced among neighboring frames in order to impose temporal consistency for estimating poses $\mathbb{P} = (p^1,p^2,...,p^T)$. The objective score function of the entire slice with temporal constrains is then given by:
\begin{equation}
\begin{aligned}
S(\mathbb{I,P})_{slice} &= \sum_{t=1}^{T}S(I^t,p^t)+\sum_{(i,i^{\ast}) \in E_f}\psi_{i,i^{\ast}}(p_{i}, p^{\prime}_{i^{\ast}}).\\
\end{aligned}
\label{equation:scoreSlice}
\end{equation}

Here $S(I^t,p^t)$ is the score function for each frame as defined in \refequ{equation:scorePose}. The pairwise term $\psi_{i,i^{\ast}}(p_{i}, p^{\prime}_{i^{\ast}})$ regularizes the temporal consistency of the part $i$ in neighboring frames. Specifically, here $p^{\prime}_{i^{\ast}} =  p_{i^{\ast}}+f_{i^{\ast},i}(p_{i^{\ast}})$ and $f_{i^{\ast},i}(p_{i^{\ast}})$ is the optical flow evaluated at $p_{i^{\ast}}$. This is the flow warping process in which pixel-wise flow tracks  are applied to align confidence values in neighboring frames to the target frame. We use the same quadratic spring model to penalize the estimation drift between these neighboring frames.\\

\noindent\textbf{Inference}\\
\label{subsection:message passing}
\noindent Inference corresponds to maximizing $S_{slice}$ defined in~\refequ{equation:scoreSlice} over $p$
for the image sequence slice.  When the relational graph $G = (V,E)$ is a tree-structured graph, exact belief propagation can be applied efficiently by one pass of dynamic programming in polynomial time. However, for cases in which the factor graph is not tree-structured but contains cycles, the belief propagation algorithm is not applicable as no leaf-to-root order can be established.
However, loopy belief propagation algorithms such as the Max-Sum algorithm make approximate inference
possible in intractable loopy models~\cite{NIPS1997_1467}. Empirical performance has consistently been reported to be excellent across various problems ~\cite{yang2016end,sigal2006measure}.
More precisely, in our case at each iteration a part $i$ sends a message to its neighbors and also receives reciprocal messages along the edges in $G$:
\begin{equation}
\begin{aligned}
score_i(p_i) &\leftarrow \phi_i(p_i|I)+\sum_{k\in child(i)}m_{ki}(p_i),
\end{aligned}
\label{equation:MP}
\end{equation}
where $child(i)$ is defined as the set of children of part $i$. The local $score_i(p_i)$ is the sum of the unary terms $\phi_i(p_i|I)$ and the messages collected from its all children. The messages $m_{ki}(p_i)$ sent from body part $k$ to part $i$ are given by:
\begin{equation}
\begin{aligned}
m_{ki}(p_i) &\leftarrow \max_{p_k}(score_k(p_k)+\psi_{k,i}(p_k, p_i))
\end{aligned}
\label{equation:MP2}
\end{equation}

\refequ{equation:MP2} computes for every location of part $i$ the best scoring location of its child $k$, based on the score of part $k$ and the spring model between $i$ and $k$. This cost maximization process can be efficiently solved via the generalized distance transforms~\cite{felzenszwalb2004distance}, reducing the computational complexity to be linear in the number of possible part locations, which is the size of the regressed heat-map from the fully ConvNet (\secref{subsection:fully ConvNet}). This inference process could be operated by several iterations till convergence. \\

\noindent\textbf{Implementation details}\\
In our implementation of the spatio-temporal message passing layer, for the first iteration, the local score for each part is initialized by its corresponding unary term obtained from the regressor layers(~\figref{fig:pipeline},(c)). The inference process is illustrated in ~\figref{fig:pipeline},(e). The children of one node could be either adjacent parts in the same frame or the same part in the neighboring frames. For the first case, the heat-maps of other parts are directly taken as input to the generalized distance transform, while for the second case the $score_k(p_k)$ is the heat-map after flow warping (\figref{fig:pipeline},(d)).
We implement message passing in a broadcasting style where messages are passed simultaneously across every edge in both directions.

Specifically, for each part $i$, \refequ{equation:MP2} computes the best score from its child $k$. The forward of this maximization process is efficiently solved via the generalized distance transform. The resulting Max location $p^*$ for each pixel is stored. Similar to the Max Pooling operation, the backpropagation of \refequ{equation:MP2} is achieved through sub-gradient decent: 
\begin{displaymath}
\frac{\partial m_{ki}(p_i)}{\partial score_k(p_k)} = \left\{ \begin{array}{cc}
1 & \mathrm{if} \ p_k = p^*, \\
0 & \mathrm{otherwise}.
\end{array}
\right.
\end{displaymath}
\begin{displaymath}
\frac{\partial m_{ki}(p_i)}{\partial \psi_{k,i}(p_k, p_i)} = \left\{ \begin{array}{cc}
1 & \mathrm{if} \ p_k = p^*, \\
0 & \mathrm{otherwise}.
\end{array}
\right.
\end{displaymath}
The gradient for the parameter of the spring model $w_{ki}$ is calculated by $\frac{\partial m_{ki}(p_i)}{\partial w_{ki}} = \frac{\partial m_{ki}(p_i)}{\partial \psi_{k,i}(p_k, p_i)}d(p_k-p_i)$, where $d(p_k-p_i)$ is the quadratic displacement. 

\section{Learning}
The learning of Thin-Slicing Network is decomposed into two stages: (1) Training fully convolutional layers and (2) Joint training with flow warping and inference layers. \\

\noindent\textbf{Training fully convolutional layers}
As discussed in \secref{subsection:fully ConvNet}, we deploy fully convolutional layers as the basic regressor to produce the belief maps for all the body parts in the sequence. As shown in \figref{fig:pipeline},(c), every pixel position has a confidence value for each joint. The ground truth heat-map for a part $i$ is written as $b_{\ast}^i(Y_i = p)$, which is created by placing a Gaussian peak at the center location of the part. In our implementation, we set peak values as $1$ and the background as $0$. We aim to minimize the $l_2$ distance between the predicted and ideal belief maps for each part, yielding the loss function:
\begin{equation}
\begin{aligned}
f = \sum_{i=1}^K\sum_{p}\left \|  b^i(p)-b_{\ast}^i(p) \right \|^2.\\
\end{aligned}
\label{equation:stage1}
\end{equation}
We use the stochastic gradient descent algorithm to train these fully convolutional layers with dropouts.\\

\noindent\textbf{Joint training with flow warping and inference layers}
For the second stage of training, the unified end-to-end model (\figref{fig:pipeline}) is jointly trained by initializing the weights of the fully convolutional layers with the pre-trained parameters. In this training stage, instead of using $l_2$ distance loss, we use the hinge loss during optimization. The final loss is defined in~\refequ{equation:hinge},
$I^i(p)$ is an indicator which is equal to 1 if the pixel lies within a circle of radius $r$ centered on the ground truth joint position, otherwise it is equal to -1:
\begin{equation}
\begin{aligned}
f = \sum_{i=1}^K\sum_{p}  \max(0,1-b^i(p)\cdot I^i(p)).\\
\end{aligned}
\label{equation:hinge}
\end{equation}
The parameters in the inference layer are differentiable so they can be end-to-end trained alongside the other weights in the  network by stochastic gradient descent.



\section{Experiments}
\noindent In this section we present results from our experimental evaluation of the proposed architecture performed on standard datasets. First we introduce the datasets and the implementation details as used during our experiments. Furthermore, we compare performance of our method with two separate baselines: a fully convolutional network and a ConvNet with spatial inference only. Finally, we compare our results with other state-of-the-art approaches across datasets.

\subsection{Datasets}
\noindent We conduct experiments on the Penn Action~\cite{Zhang2013From} and JHMDB~\cite{Jhuang:ICCV:2013} datasets, both standard datasets to evaluate video-based pose estimation.\\

\noindent\textbf{Penn Action dataset}
the Penn Action dataset~\cite{Zhang2013From} is one of the largest datasets with full annotations of human joints in videos,containing 2326 unconstrained videos depicting 15 different action categories and the annotations include 13 human joints for each image. An additional occlusion label for each joint is also provided. We follow the original paper~\cite{Zhang2013From} to split the data into training and testing subsets in a roughly half-half manner. In total there are around 90k images for training and 80k images for testing.\\

\noindent\textbf{JHMDB dataset}
The JHMDB dataset~\cite{Jhuang:ICCV:2013} contains 928 videos and 21 action classes. The dataset provides three different splits of training and testing, and we report the average performance over these three splits for all evaluations on this dataset. The experiments on a subset of this dataset (sub-JHMDB dataset) are also conducted to compare with other state-of-the-art methods. This subset contains 316 clips with 12 action categories. In this subset the whole human body is inside the image so all joints are annotated with ground truth positions.


\subsection{Implementation Details}
\noindent\textbf{Data augmentation}
to introduce more variation in the training data and thus reducing overfitting, we augment the data by rotating images between -90 to 90 degrees chosen randomly and by scaling by a random factor between 0.5 to 2. When pre-training the fully convolutional layers, the inputs to the network are the cropped image patch around the center of persons with random shifts. For end-to-end training with the flow warping and spatio-temporal message passing layer, the input patches for the sequence are controlled to have the same pre-processing.  \\

\noindent\textbf{Network parameter settings}
for the fully convolutional layers, we deploy the network structure based on~\cite{wei2016cpm}. This model has a multiple-stage structure which is designed to alleviate the problem of vanishing gradients. We use an input size of 368 $\times$ 368 px in order to cover sufficient context. The batch size is set to 20 for pre-training the convolutional layers and 6 for jointly training the unified network respectively when the thin-slicing is 5 frames. The learning rates are initialized as 0.0005 for the first stage of training and dropped by a factor of 3 every 20k iterations. For end-to-end training, the learning rate is set to be lower (0.0001) and is dropped every 5k iterations also by a factor of 3. The dropout rate is set to 0.5 for the first stage and increased to 0.7 for the second stage with flow warping and message passing layers to reduce potential effects of overfitting. The fully ConvNet is trained for 10 epochs for initialization. The unified end-to-end model typically converges after 3-4 epochs. The flow warping layer takes resized optical flow images of the same size as the heat-maps as input with their values rescaled by the same scaling factor.

For the spatio-temporal message passing layer, we initialize the weight of the quadratic term to 0.01 and the first-order term to 0 for the generalized distance transform algorithm~\cite{felzenszwalb2004distance}. Please note that setting the normalization terms when collecting messages sent from children can help stabilize the training process. A similar observation is also reported in~\cite{yang2016end}. We find that three iterations of approximate inference already provides satisfactory results and if not specified otherwise message passing is stopped after three iterations in our experiments.\\

\noindent\textbf{Edge connections in the graph}
The spatio-temporal loopy structure used in this implementation is visualized in ~\figref{fig:pipeline}, (e). Spatially, the
structured model has edges coinciding with body limbs and it additionally connects symmetric body parts (e.g., left wrist and right wrist, left knee and right knee) to alleviate the double counting issue. Temporal edges connect the same body parts across two adjacent frames. However, our implementation of the inference layer is flexible and can perform approximate inference on arbitrary loopy graph configurations.

\subsection{Evaluation Protocol}
For consistent comparison with prior work on both the Penn Action dataset and the JHMDB dataset~\cite{gkioxari2016chained,xiaohan2015joint,park2011n}, we use a metric referred to as PCK, introduced in ~\cite{yang2011articulated}. A candidate keypoint prediction is considered to be correct if it falls within $\alpha \cdot \max(h, w)$ pixels of the
ground-truth keypoint, where $h$ and $w$ are the height
and width of the bounding box of the instance in question, and $\alpha$ controls the relative threshold for considering correctness. We report results from different settings of $\alpha$.
We also report results that plot accuracy vs normalized distance from ground truth in pixels, where a joint is deemed correctly located if it is within a set distance
of $d$ pixels from a ground-truth joint center, where $d$ is normalized by the size of the instance.

\subsection{Result Analysis for Penn Action Dataset}
\noindent\textbf{Baseline comparison:} \reftab{table:Penn} shows the relative performance on the Penn Action test set. For consistent comparison with previous work~\cite{xiaohan2015joint,gkioxari2016chained,park2011n}, the metric PCK@0.2 is used, which means a prediction is considered correct if it lies within $(\alpha=0.2) \times \max(s_h,s_w)$. We first compare the results from the pure ConvNet baseline model, the spatial-only model and finally our spatio-temporal inference model. The baseline model corresponds to the pure fully ConvNet as described in~\secref{subsection:fully ConvNet} and is trained with loss~\refequ{equation:stage1}. We also report the result after only applying spatial inference on top of the heat-maps obtained from the ConvNet, coresponding to only the blue arrows  in~\figref{fig:pipeline},(e). Please note that these two settings essentially treat video-based pose estimation as pure concatenation of single image predictions. Finally, we report the performance of our proposed end-to-end trainable network with full spatio-temporal inference. \\
\begin{table}[t]
\begin{center}
\resizebox{\linewidth}{!}{
\begin{tabular}{l|cccccccc}
\hline
 Method& Head & Shou &Elbo &Wris &Hip &Knee &Ankl  & Mean \\
\hline
\cite{park2011n} &  62.8 & 52.0 & 32.3 & 23.3& 53.3& 50.2& 43.0& 45.3 \\
\cite{xiaohan2015joint} &   64.2 & 55.4 & 33.8 & 24.4& 56.4& 54.1& 48.0& 48.0 \\
\cite{iqbal2016pose} &   89.1 & 86.4 & 73.9 & 73.0& 85.3& 79.9& 80.3& 81.1 \\
\cite{gkioxari2016chained} & 95.6 & 93.8 & 90.4 & 90.7& 91.8& 90.8& 91.5& 91.8 \\
\hline
baseline& 97.9 &94.9 &76.8 &72.0 &95.9 &88.8 &85.1 & 87.0\\
S-infer & 98.0 &90.3 & 85.2 &86.7 & 93.7& 93.5 & 93.6 &91.4\\
ST-infer & 98.0 & 97.3 & 95.1 & 94.7 &97.1 &97.1 & 96.9 &96.5\\
\hline
ST-infer($\star$) & 97.9 & 91.1 & 91.3 & 90.9 &92.5 &94.4 & 94.5 &92.8\\
ST-infer($\ast$) & 97.9 & 89.7 & 84.4 & 86.5 &93.4 &93.7 & 93.8 &91.0\\
ST-infer($2$) & 97.6 & 96.8 & 95.2 & 95.1 &97.0 &96.8 & 96.9 &96.4\\
\hline
\end{tabular}
}
\end{center}
\caption{Comparison of PCK@0.2 on Penn Action dataset. We compare our proposed model with baseline model, baseline model with spatial inference and other state-of-the-art methods. We also investigate the performance of independent training ($\star$), the baseline ConvNet after end-to-end training ($\ast$) and temporal connection across 2 frames ($2$). }
\label{table:Penn}
\end{table}
Our baseline setting achieves $87.0\%$ average accuracy for all 13 body parts. Spatial inference with geometric constraints among human body parts in individual images increases the overall result by $4.4\%$. By incorporating temporal consistency across frames, we observe an additional accuracy gain of $5.1\%$ over spatial inference only.\\
\begin{figure*}[h]
\begin{center}
\subfigure{
\label{fig:objdetection}
\includegraphics[width=0.25\linewidth]{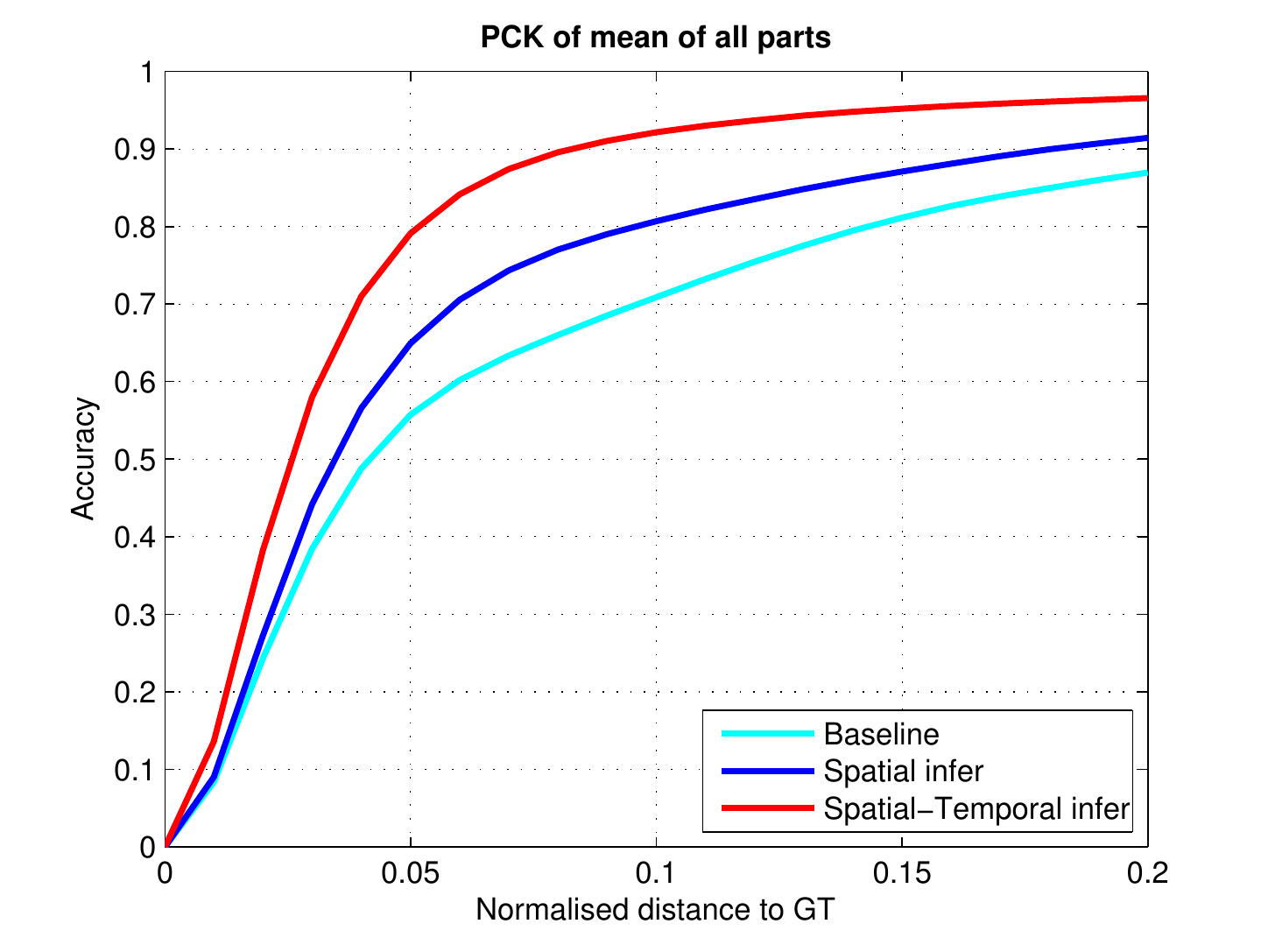}
}\hspace*{-1.5em}
\subfigure{
\label{fig:objdetection}
\includegraphics[width=0.25\linewidth]{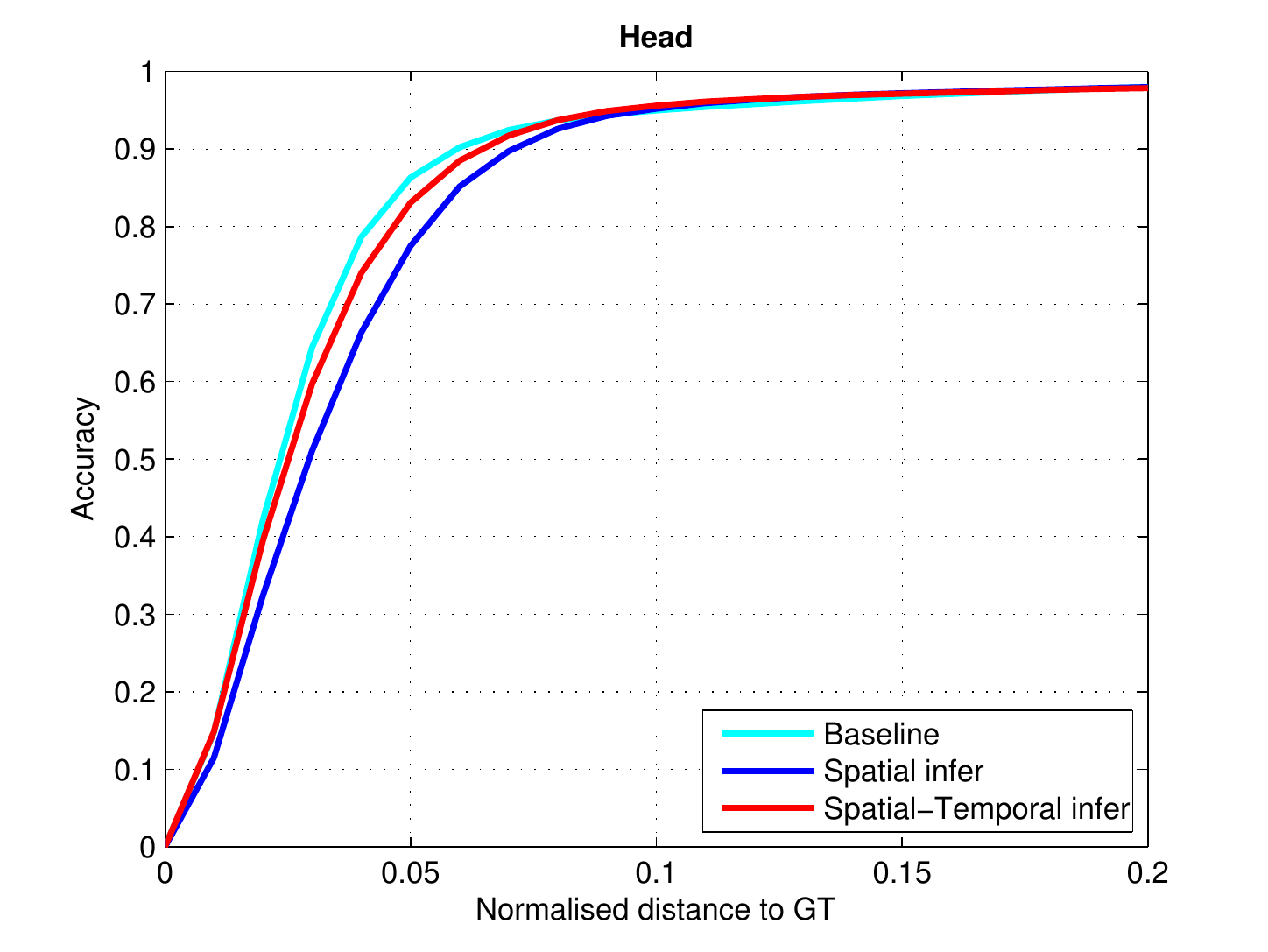}
}
\subfigure{
\label{fig:objdetection}
\includegraphics[width=0.25\linewidth]{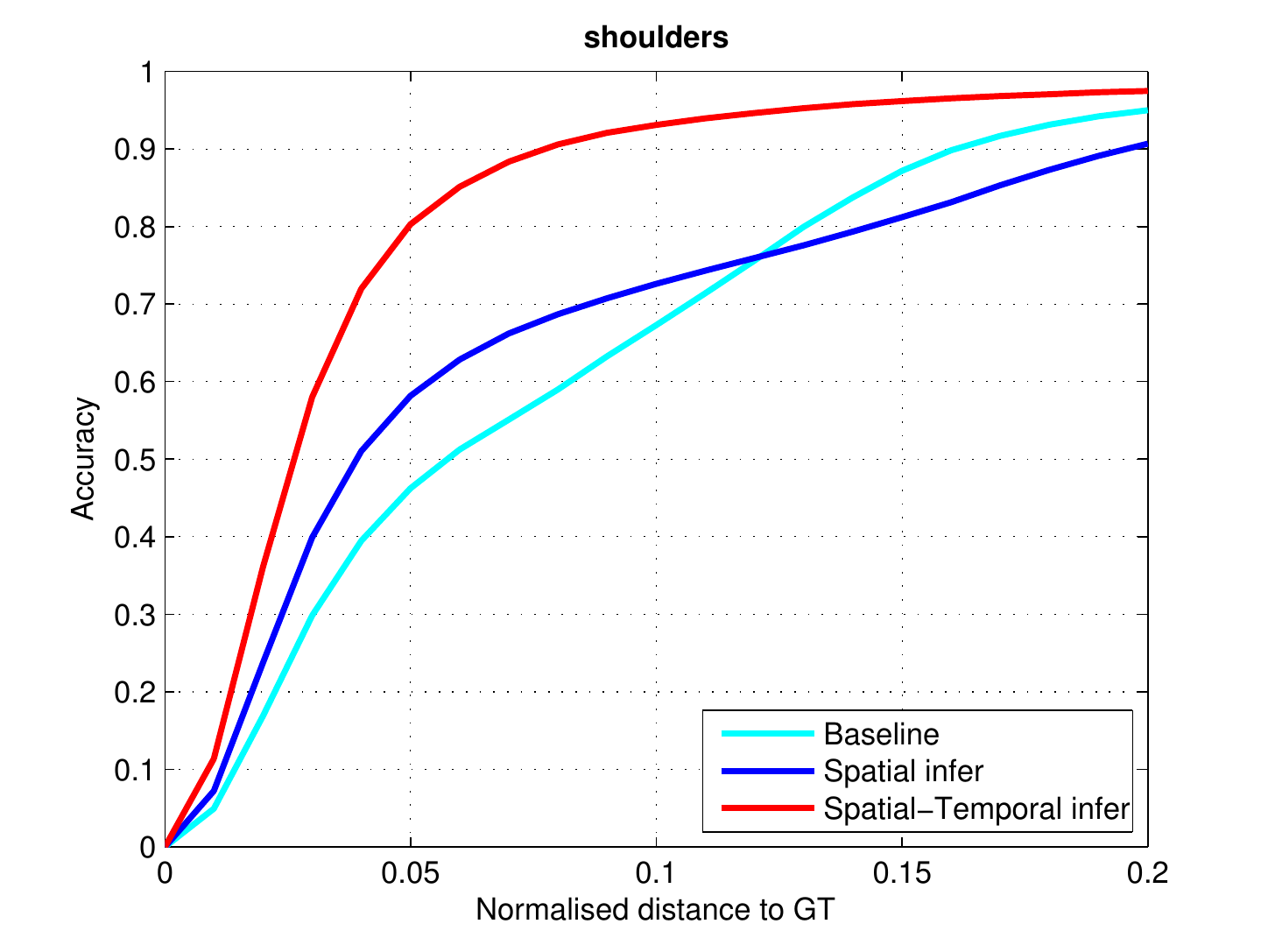}
}\hspace*{-1.5em}
\subfigure{
\label{fig:objdetection}
\includegraphics[width=0.25\linewidth]{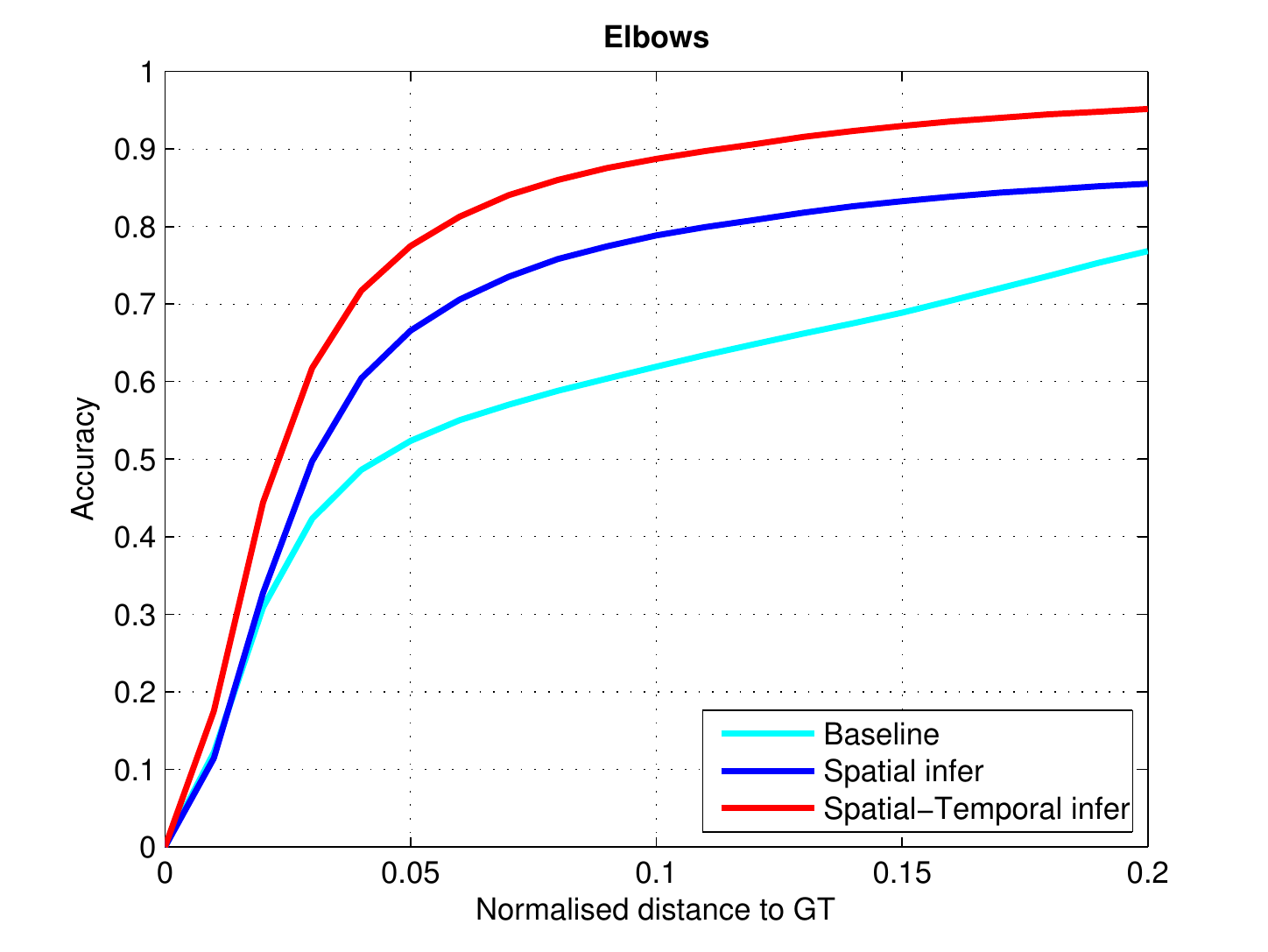}
}
\vspace{-3.5mm}
\subfigure{
\label{fig:objdetection}
\includegraphics[width=0.25\linewidth]{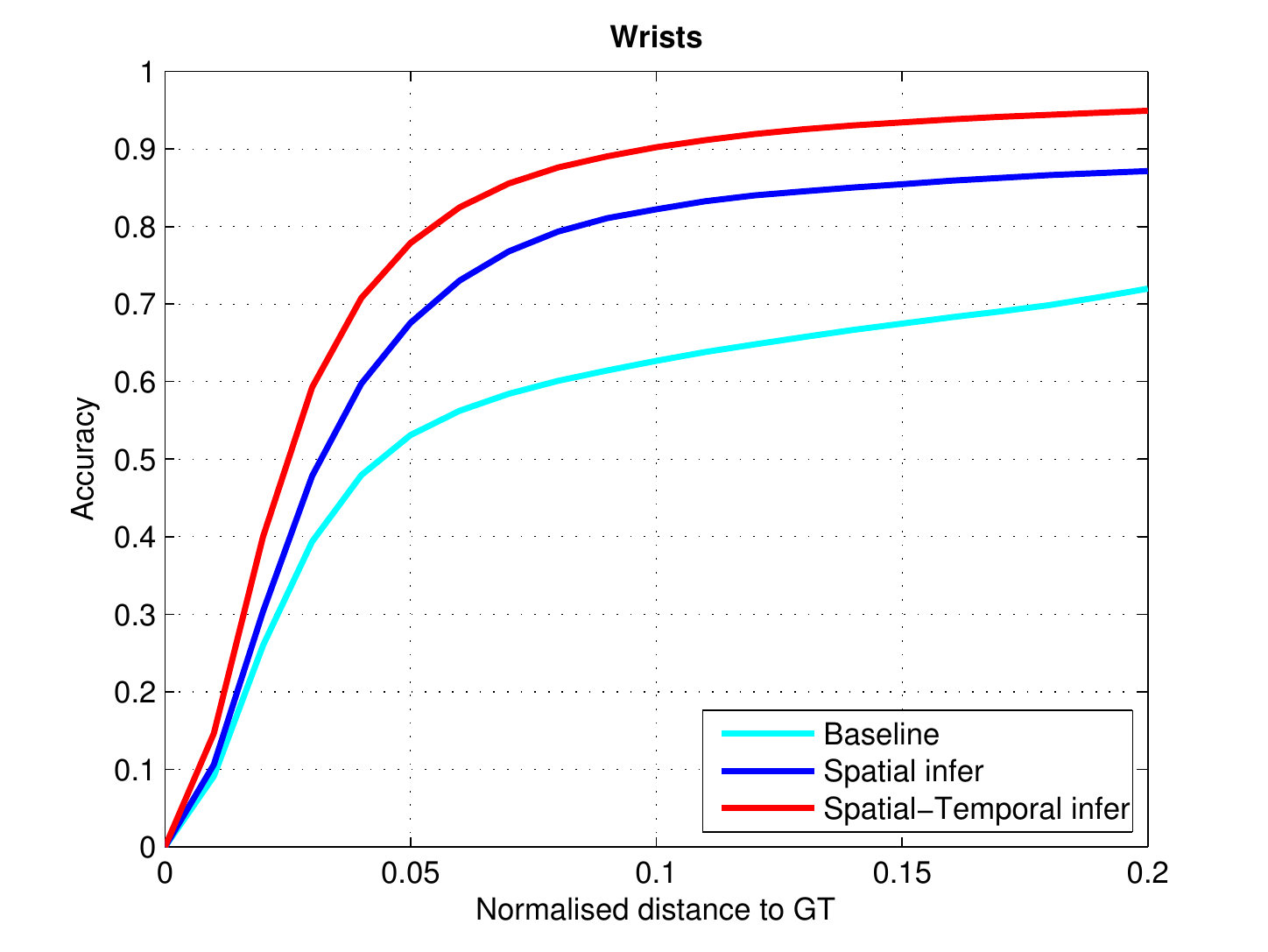}
}\hspace*{-1.5em}
\subfigure{
\label{fig:objdetection}
\includegraphics[width=0.25\linewidth]{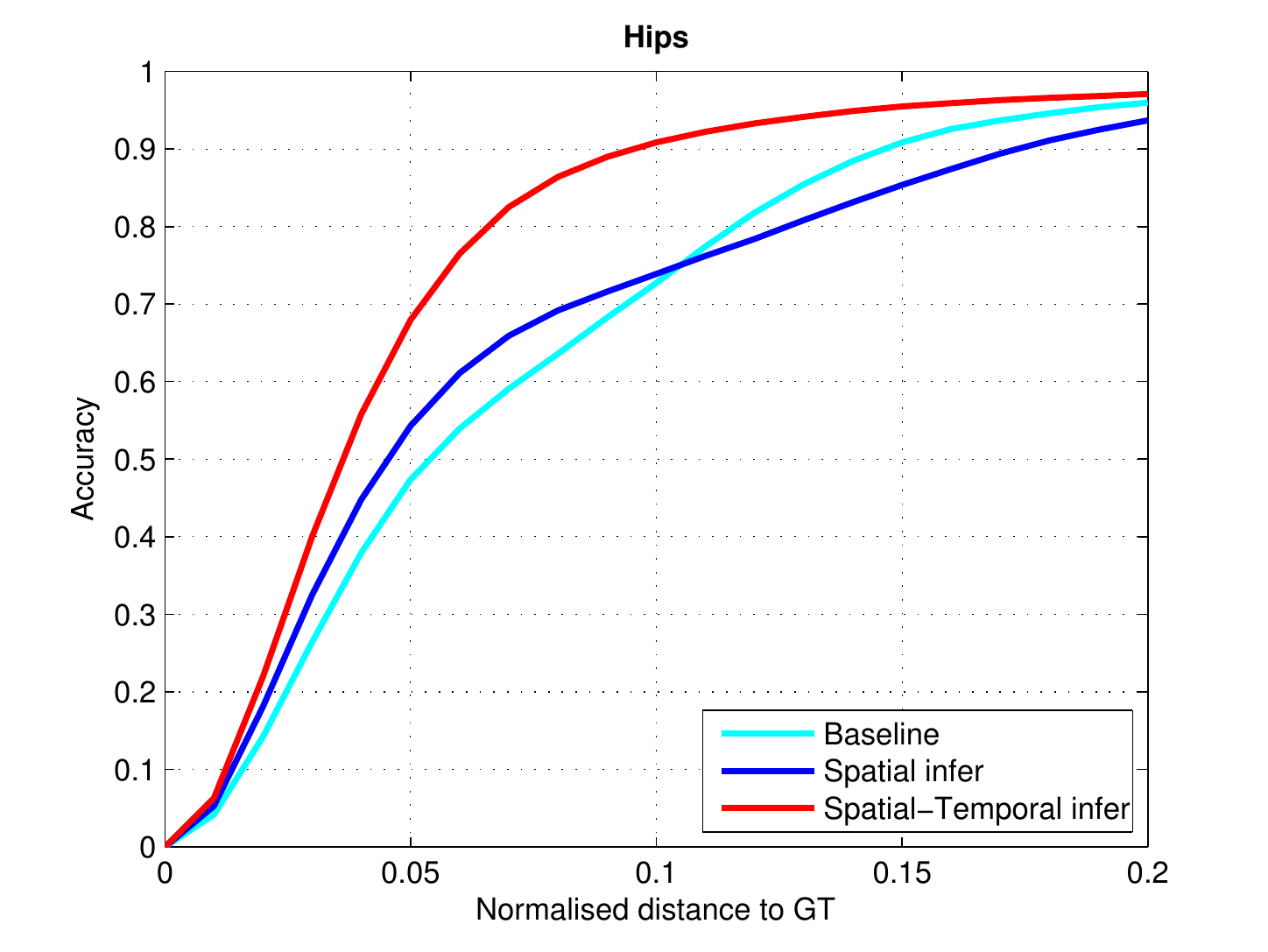}
}%
\subfigure{
\label{fig:objdetection}
\includegraphics[width=0.25\linewidth]{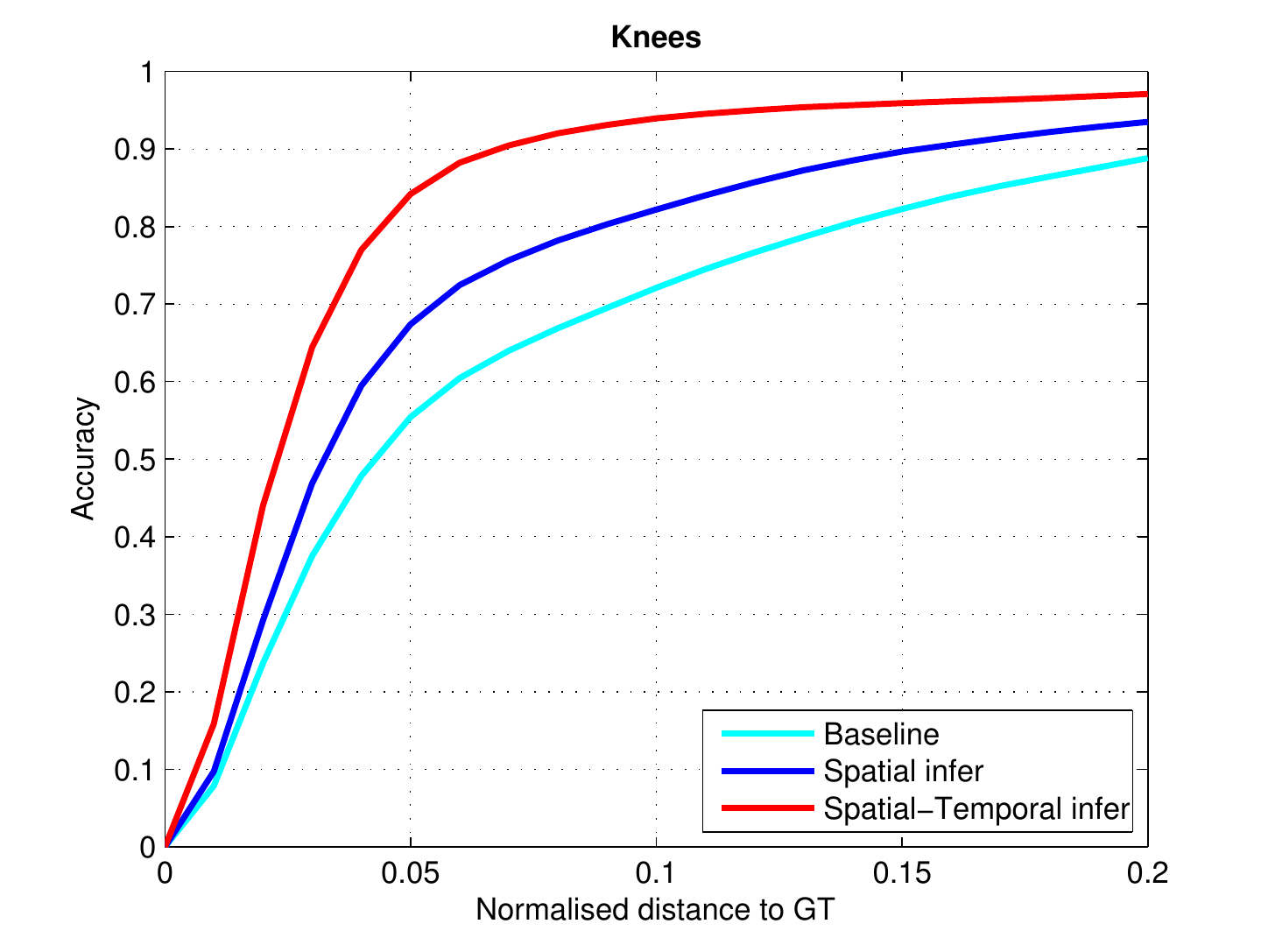}
}\hspace*{-1.5em}
\subfigure{
\label{fig:objdetection}
\includegraphics[width=0.25\linewidth]{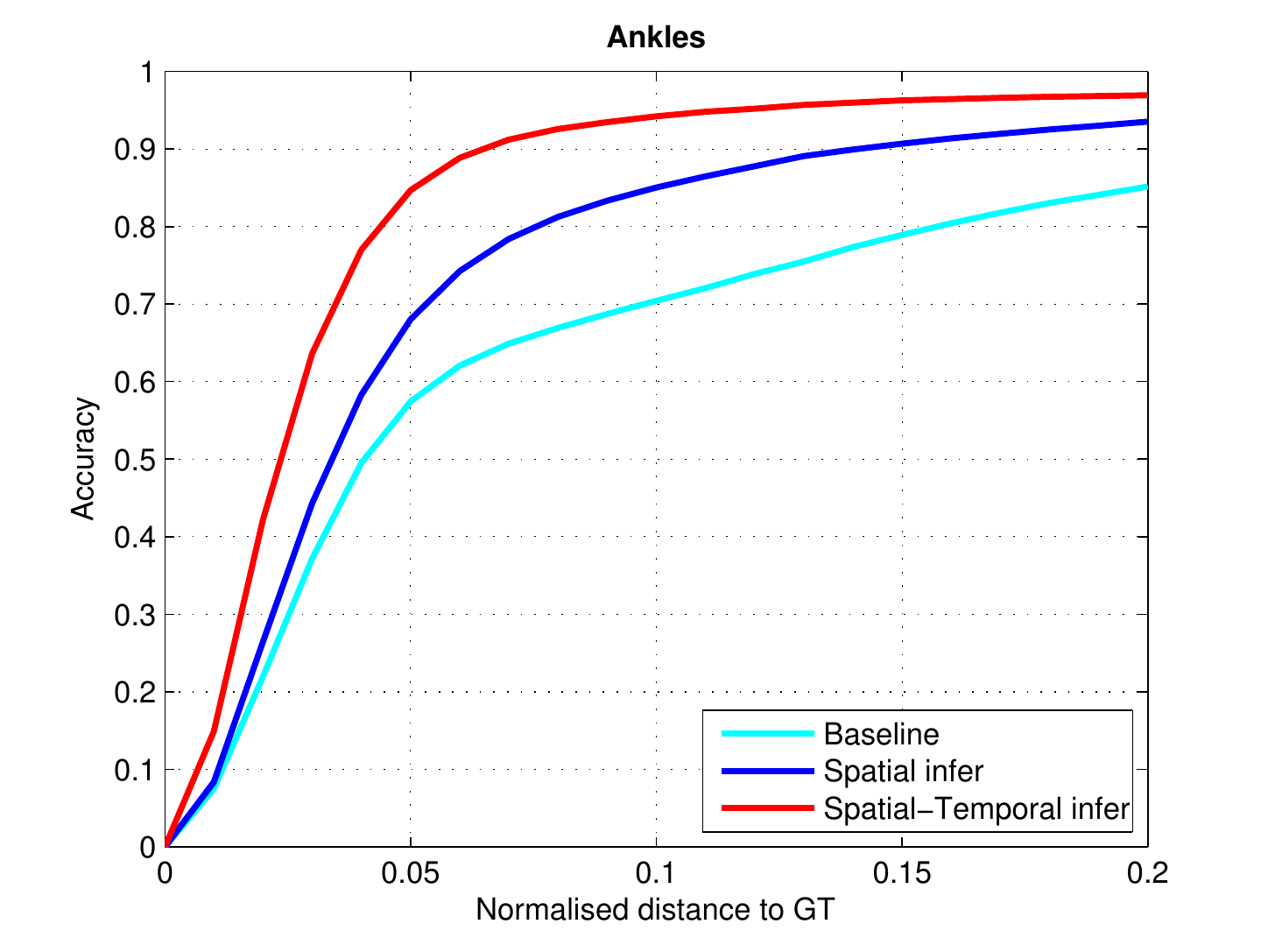}
}%
\end{center}
   \caption{PCK curve for Penn Action dataset. We compare our proposed model with two baselines -- ConvNet-only and spatial inference-only. Ours yields consistent accuracy improvements across the entire range of strictness.}
\label{fig:PCK curve Penn}
\end{figure*}
Body parts like head and shoulders are usually visible and less flexible, so even with the baseline model very high detection accuracies can be achieved. However, parts such as elbows and wrists are the most flexible joints of our body. This flexibility can yield configurations with very large variation and these joints are also prone to be occluded by other parts of the body. This is shown by the low detection rates from the baseline part-regression model. With spatial message passing, the accuracy increases, and our proposed model boosts this again by roughly $10\%$. Note that predictions for shoulders can be negatively influenced by sending or receiving messages from elbows through spatial inference only. However, deploying temporal information helps in recovering from such errors.

\noindent\textbf{Analysis of normalized distance curves}
~\figref{fig:PCK curve Penn} plots the normalized distance to the ground truth annotations. Generally, our proposed model outperforms the baseline model and the one with spatial inference over all levels of the evaluation and across all joints. Interestingly, even for stable (and hence easy to predict) joints like the head, we can still see improvements. In particular when the metric gets more strict (i.e., smaller $d$). In the cases of more flexible pody parts such as elbows, wrists and knees, a constant improvement for both loose and strict metric can be observed. Especially over the 0.05 to 0.1 region, we can clearly observe more accurate predictions. This further suggests that back-propagating the error from several frames through our spatio-temporal network architecture benefits both unary and pairwise terms.

\noindent\textbf{Further evaluations}
We also test the effectiveness of joint training of convolutional layers with message passing. Keeping the weights of convolutional layers fixed, we just train the parameters in the spatio-temporal inference layer. The overall performance is $92.8\%$ (Table~\ref{table:Penn}, row annotated by ($\star$)). It improves over the baseline model by $5.8\%$ but could not reach the performance of joint training.
As mentioned previously, the end-to-end training helps the fully convolutional layers to capture appearance features better. To validate this claim we conduct the same evaluation using the convolutional layers from the end-to-end trained model (removing the spatio-temporal inference layers) and compare the result with the baseline model (trained standalone). An overall $4\%$ performance increase (Table~\ref{table:Penn}, row annotated by ($\ast$)) can be observed.
We also perform the experiment with temporal edges across not only 1 frame but 2 frames (Table~\ref{table:Penn}, row annotated by (2)). However, here we do not observe a significant increase of mean accuracy.

\noindent\textbf{Comparison with state-of-the-art}
Table~\ref{table:Penn} also lists the comparison between the results of
previous methods and ours. We first compare with shallow hand-crafted features based works~\cite{xiaohan2015joint,park2011n}. \cite{park2011n} is based on N-best algorithm and ~\cite{xiaohan2015joint} employs different action specific models. We use the figures reported in \cite{xiaohan2015joint} for comparison. We outperform them by a large margin for all body parts. \cite{gkioxari2016chained} incorporates deep features with a recurrent structure to model long-term dependency between frames. While only propagating information over short periods of time (thin-slices of the sequence), we still attain an overall performance boost of $4.7\%$ on this dataset. Please note that ours consistently localizes all joints better than prior work.


\begin{table}[t]
\begin{center}
\resizebox{\linewidth}{!}{
\begin{tabular}{l|cccccccc}
\hline
 Method& Head & Shou &Elbo &Wris &Hip &Knee &Ankl &Mean \\
\hline
baseline & 93.2 & 72.4 &57.3 &61.9 &88.4 &63.6 &48.6& 70.9 \\
S-infer & 93.6 & 85.1 &72.9 &70.1 &87.2 &66.2 &52.2& 76.5\\
ST-infer & 93.6 & 94.7 &84.8 &80.2 &87.7 &68.8 &55.2& 81.6\\
\hline
baseline($\ast$) & 86.2 & 50.2 &42.9 &47.4 &61.4 &43.4 &34.1& 54.5 \\
S-infer($\ast$) & 86.1 & 62.8 &55.2 &51.9 &68.3 &48.1 &36.7& 60.2 \\
ST-infer($\ast$) & 85.4 & 77.6 &69.4 &62.6 &76.9 &57.4 &42.9& 68.7 \\
\hline
\end{tabular}
}
\end{center}
\caption{Results on full JHMDB dataset. The first three rows are based on PCK@0.2 while the results with ($\ast$) are with PCK@0.1.}
\label{table:JHMDB}
\end{table}

\begin{figure*}[h]
\begin{center}
   \includegraphics[width=0.95\linewidth]{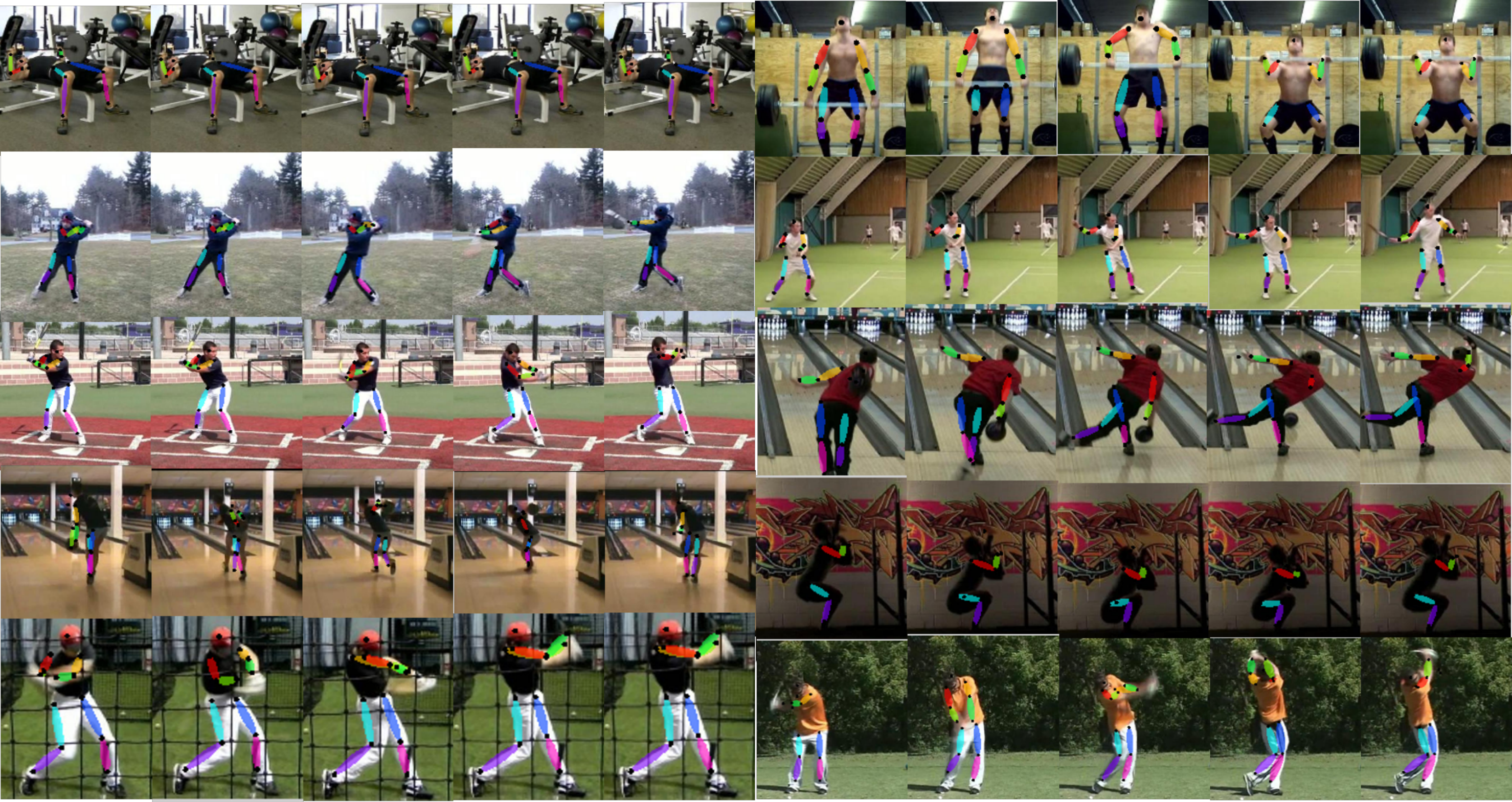}
\end{center}
   \caption{Qualitative results on Penn Action dataset. We visualize
connections among challenging limbs (arms and legs). Some failure cases are listed. Our method
may miss limbs due to significant occlusions and heavy blur (last row).}
\label{fig:sample_sequence}
\end{figure*}

\subsection{Result Analysis for JHMDB dataset}
We also conduct a systematic evaluation on the JHMDB dataset~\cite{Jhuang:ICCV:2013}. The average result of three splits on this dataset is illustrated in \reftab{table:JHMDB}. The first three rows summarize the performance under the PCK@0.2 metric. The same three models and settings as previously are evaluated and we observe results consistent with the experiments conducted on the Penn Action dataset. The proposed end-to-end model boosts the overall performance by a relatively large margin. We also provide results for PCK@0.1 (\reftab{table:JHMDB}, row marked with $\ast$).
\begin{table}[h]
\begin{center}
\resizebox{\linewidth}{!}{
\begin{tabular}{l|cccccccc}
\hline
 Method& Head & Shou &Elbo &Wris &Hip &Knee &Ankl &Mean \\
\hline
\cite{park2011n}  & 79.0 & 60.3 &28.7  &16.0   &74.8 &59.2  &49.3& 52.5 \\
\cite{xiaohan2015joint} & 80.3 & 63.5 &32.5  &21.6   &76.3 &62.7  &53.1& 55.7 \\
\cite{iqbal2016pose}& 90.3 & 76.9 &59.3  &55.0   &85.9 &76.4  &73.0& 73.8 \\
\hline
baseline & 97.2 & 82.2 &65.2  &66.5   &96.3 &84.4  &76.8& 82.3 \\
S-infer & 97.0 & 87.3 &74.9 &71.1 &97.5 &89.4 &86.0& 86.9\\
ST-infer & 97.1 & 95.7 &87.5 &81.6 &98.0 &92.7 &89.8& 92.1\\
\hline
\end{tabular}
}
\end{center}
\caption{PCK@0.2 results on sub-JHMDB dataset. We compare with other previous methods and our own baselines.}
\label{table:sub-JHMDB}
\end{table}
To consistently compare with other state-of-the-art results, we perform further experiments on a subset of the JHMDB dataset. These subsets remove sequences with incomplete bodies. The comparison is listed in \reftab{table:sub-JHMDB}. We outperform shallow feature based methods by a large gap \cite{park2011n,xiaohan2015joint}. In \cite{iqbal2016pose}, features are taken from the deep ConvNet and a graphical model based inference is conducted independently to refine the result. Our proposed method also provides better performance across all body parts.

\subsection{Qualitative results}
\figref{fig:sample_sequence} illustrates results from representative sequences taken from our experiments. Our method can capture articulated poses with strong pose changes across several frames. Cases with cluttered background, occlusion, and blur are included. Failure cases, shown in the bottom row of \figref{fig:sample_sequence}, are often linked to extended periods of motion blur or occlusion across frames. This hinders the ConvNet from capturing local appearance properties and impacts the estimation of dense optical flow. In these cases temporal inference over longer distances may be necessary.



\section{Conclusion}\label{sec:conclusion}
We have proposed an end-to-end trainable network taking spatio-temporal consistency into consideration to estimate human poses in natural, unconstrained video sequence. We have experimentally shown that leveraging such a unified structured prediction approach outperforms multiple baselines and state-of-the art methods across datasets. Training regression layers jointly with the spatio-temporal inference layer benefits cases that display motion blur and occlusions but also improves predictions of unary terms due to the iterative back-propagation of errors. Interesting directions for future work include incorporation of long-range temporal dependencies and handling of groups of people.

\section*{Acknowledgement}
This work is partially supported by the ERC Advanced Grant VarCity.


{\small
\bibliographystyle{ieee}
\bibliography{egbib}
}

\end{document}